\documentclass[conference]{IEEEtran}
\IEEEoverridecommandlockouts
\usepackage{cite}
\usepackage{amsmath,amssymb,amsfonts}
\usepackage{algorithmic}
\usepackage{graphicx}
\usepackage{textcomp}
\usepackage{xcolor}
\def\BibTeX{{\rm B\kern-.05em{\sc i\kern-.025em b}\kern-.08em
    T\kern-.1667em\lower.7ex\hbox{E}\kern-.125emX}}
\begin{document}
\title{Comparative Studies of Unsupervised and Supervised Learning Methods based on Multimedia Applications\\
}

%}

\author{\IEEEauthorblockN{1\textsuperscript{st} Amitesh Kumar Singam}
\IEEEauthorblockA{\textit{ IEEE Computational Intelligence Society} \\
\textit{Institute of Electrical and Electronics Engineers}\\
Hyderabad, India \\
0000-0002-2532-0989}
\and
\IEEEauthorblockN{2\textsuperscript{nd}Benny Lövström}
\IEEEauthorblockA{\textit{Blekinge Institute of Technology}\\
Karlskrona, Sweden \\
0000-0003-3824-0942}\and
\IEEEauthorblockN{3\textsuperscript{rd}Wlodek J. Kulesza}
\IEEEauthorblockA{\textit{Blekinge Institute of Technology}\\
Karlskrona, Sweden  \\
0000-0001-8351-8830}
}
\maketitle
\begin{abstract}
%\boldmath
In the mobile communication field, some of the video applications boosted the interest of robust methods for video quality assessment. Out of all existing methods, We Preferred, No Reference Video Quality Assessment is the one which is most needed in situations where the handiness of reference video is partially available. Our research interest lies in formulating and melding effective features into one model based on human visualizing characteristics. Our work explores comparative study between Supervised and unsupervised learning methods. Therefore, we implemented support vector regression algorithm as NR-based Video Quality Metric(VQM) for quality estimation with simplified input features. We concluded that our proposed model exhibited sparseness even after dimension reduction for objective scores of SSIM quality metric.
\end{abstract}
\begin{IEEEkeywords}
VQM, NR-VQM, SVM, PCA, SSIM.
\end{IEEEkeywords}

% \fi
\IEEEpeerreviewmaketitle

\section{Introduction}
The usage of consumer video applications in mobile devices has been increased to a large extent. Due to the extremely large competition between service providers and application developers to provide the better quality based on the neediness for the advanced methods to assess the video quality is in great demand now. The Multimedia services like video chatting and live video streaming in mobile or hand-held devices boosted the interest in no-reference objective video quality assessment since the availability of reference video is partially available. One of the main features of video services is quality of service as observed by the end user. For instance, the visual quality within transmission can get degraded while capturing, loading, saving or reproduction and distortions within video buffering might occur at any of these stages. Since humans are considered as true judges of video quality due to end users of the video services. In technical terms, the process of validating of video quality by viewers is referred as subjective video quality assessment. However, subjective experiments are often too time-consuming, inconvenient, costly and moreover it must be done under specific recommendations in order to produce standard results. These facts gave rise to the need of some intelligent ways of  predicting  quality assessment based on accuracy and it can be performed swiftly and economically.
\section{Video Quality Assessment}
 In our research work, we carefully employed the specifications recommended by ITU-R BT 500-10 as mentioned in\cite{1} and VQEG Hybrid test plan in\cite{6} which is explained briefly in following sections.
\section{Proposed Idea}
Mainly, the proposed method involves extraction of visual quality relevant bitstream parameters and building of a machine learning based model for quality prediction. These parameters were selected carefully to keep the complexity in control and to get reasonable coding information which can represent the coding distortions. Following is the description of the extracted parameters and the rationale of making a parameter a part of the proposed model.
\section{Feature Extraction of H.264 Bitstream Data}
Feature extraction out of encoded video sequence has been processed using JM Reference software Version16.0. The Feature extraction of bit stream data which has been generated as a trace file after encoding process.
\label{features}
\begin{enumerate}
\item   Bitrate.
\item	Frame rate
\item	Percentage of inter macroblocks of size16x16.
\item	Percentage of inter macroblocks of size 4x4
\item	Percentage of inter macroblocks of size 8x8
\item  Average quantization parameter.
.
\end{enumerate}
\section{Unsupervised Learning Method}
Generally when we are dealing with high dimensional data, addition of more features will effect the system or model performance and increase complexity of system, therefore in order to overcome curse of dimensionality factors such as efficiency, classification performance and ease of modeling are to be considered. Efficiency is further classified into measurement, storage and computational costs. Especially in machine learning concept, a finite number of observations in a high-dimensional induced space with each input data having possible number of values, any model required large number of features for training to make sure that they are numerous samples with each permutation of values. Therefore, our main idea is to reduce the dimension of inputs with minimal loss of information. Dimensionality reduction is suitable in visualizing data, noticing a compact representation, and minimizes computational processing. In addition, reducing the number of dimensions can separate the features with significant data from less significant ones which provides further Vision into the nature of the data which may not be discovered otherwise. There are various dimensionality reduction techniques like Partial least squares(PLS), principle component analysis(PCA), singular value decomposition(SVD), kernel principle component analysis(KPCA), factor analysis and hierarchical clustering etc. PLS and PCA has two functions in regression analysis i.e. transforming highly correlated variables to independent variables via linear transformation and dimensionality reduction. In the case of regression, So we preferred to use PCA for dimensionality reduction in our research work.
\subsection{Dimension Reduction}
\paragraph{}Especially in machine learning concept, a finite number of observations in a high-dimensional induced space with each input data having a large number of possible values, any model requires large number of features for training to make sure that they are numerous samples with each permutation of values. Therefore, our main idea is to reduce the dimension of inputs with minimal loss of information. Dimensionality reduction is suitable in visualizing data, noticing a compact representation, and minimizes computational load. In addition, reducing the number of dimensions can separate the features with significant data from less significant ones which provides further vision into the nature of the data which may not be discovered otherwise. There are various dimension reduction techniques like principle component analysis, singular value decomposition, kernel principle component analysis, factor analysis and hierarchical clustering etc. In our research work we deployed PCA for dimensionality reduction
\subsection{Principle Component Analysis }
Principle component analysis is a statistical method which uses orthogonal linear transformation technique that projects a set of vectors into a new dimension which has components that are linearly uncorrelated and arranged according to decreasing order of variance. It is assumed that most significant data is found in the first coordinates of the projected space and it contains larger variance. Reducing the dimension of the input vector leads to improvement in the generalization performance. 
\paragraph{}PCA is used to reduce the dimension of input space for our proposed model. The reduction of dimensions in the input space will reduce the complexity of the system and also decreases the time which is necessary to train the model. The PCA technique can be chosen as the method for preprocessing data and to extract uncorrelated features from the data.
\paragraph{}Computing PCA, If $\vec{X}$ is an input vector of m dimensions and In order to reduce the dimensions, we need to eliminate insignificant data from a given input vector.
\begin{equation}
\vec{X}=[x_1,x_2,..,x_{m0},x_{m0+1},.....,x_m]
\label{equation:eq8}
\end{equation}
\paragraph{} we have to map the input vector $\vec{X}$ in other dimension$ \vec{X}_{new}$. By designing a transformation matrix (T) of m rows and m columns and multiplying matrix T with input vector we get$ \vec{X}_{new}$ . Therefore expression is given as
\begin{equation}
\vec{X}_{new}=T.\vec{X}
\end{equation}
\begin{itemize}
\item X is a zero mean m dimensional random vector.
\item$\vec{X}_{new}$ is new space dimension.
\end{itemize}
\paragraph{}PCA is a multivariate technique employed for dimensionality reduction of a Input features with high number of correlated variables, Therefore we need to maximize the rate of decrease of variance in order to reduce the dimensionality of input vector. A mathematical model of PCA follows
\paragraph{}If $\vec{X}$ is input random vector and q is unit vector of m dimension and According to the property of vector $\vec{q}$, Euclidean norm of $\vec{q}$ is one. $ \parallel \vec{q}\parallel=1$
By projecting X on q in a projection space A. we get,
\begin{equation}
A=\vec{X}^T\vec{q}
\end{equation}
Variance
\begin{equation}
E[A^2]=\vec{q}R_{xx}\vec{q}{^T}
\label{equation:eq9}
\end{equation}
where $R_{xx}$ is the correlation matrix
\begin{equation}
R_{xx}=E[\vec{X}\vec{X}{^T}]
\end{equation}
let $\psi(\vec{q})$ be the variance probe
\begin{equation}
\psi(\vec{q})=\vec{q}R_{xx}\vec{q}{^T}
\label{equation:eq11}
\end{equation}
For minimal value of variance
\begin{equation}
\psi(\vec{q}+\delta \vec{q})=\psi(\vec{q})
\label{equation:eq12}
\end{equation}
the eigen values $\lambda_1, \lambda_2,.......... \lambda_m$ and the corresponding orthogonal eigenvectors $\vec{q}_1,\vec{q}_2,......\vec{q}_m$ of the covariance matrix R are calculated and arranged according to their magnitude $\lambda_1\geq\lambda_2\geq.......... \geq\lambda_m$
\begin{equation}
R\vec{q}_j=\lambda_j\vec{q}_j
\end{equation}
where$\lambda_1$ is highest value of all eigen values.
\begin{equation}
\vec{Q}=[\vec{q}_1,\vec{q}_2,...........\vec{q}_j,........,\vec{q}_m]
\label{equation:eq13}
\end{equation}
\begin{equation}
\Delta=diagonal[\lambda_1, \lambda_2,.......... \lambda_m]
\label{equation:eq14}
\end{equation}
$\vec{Q}$ is an orthogonal matrix satisfying
\begin{equation}
\vec{q}_i{^T}\vec{q}_j= \left.
 \begin{array}{l l}
1, \text{if $i=j$ }\\
0, \text{if $i\neq j$ }\\
\end{array} \right.
\label{equation:eq15}
\end{equation}
Orthogonal similarity transformation is expressed as
\begin{equation}
\vec{Q}^TR\vec{Q}=\Delta
\end{equation}
In expanded form
\begin{equation}
 \vec{q}_j{^T} R \vec{q}_k= \left.
\begin{array}{l l}
\lambda_j, \text{if $k=j$ }\\
 0, \text{if $k\neq j$ }\\
\end{array} \right.
\label{equation:eq16}
\end{equation}
The correlation matrix R is expressed in terms of its eigenvalues and eigenvectors as following
\begin{equation}
 R=\sum_{j=1}^m \lambda_j \vec{q}_j[\vec{q}]_j{^T}
\end{equation}
\begin{equation}
\psi(\vec{q}_j)=\lambda_j
\end{equation}
\begin{equation}
a_j=X{^T} \vec{q}_j
 \end{equation}
for j=1,2,....m (represents analysis)
 $ a_j$=projection of X onto principle directions,
\begin{equation}
{X}=\sum_{j=1}^m a_j \vec{q}_j
\end{equation}
 for j=1,...., m       (represents synthesis)
Let the eigen values $\lambda_1,\lambda_2,.....\lambda_l $  are largest values of correlation matrix $R_{xx}$  and
\begin{equation}
\hat{X}=\sum_{j=1}^l a_j \vec{q}_j
\end{equation}
$\hat{X}$ is not exact but approximate solution because of truncation to l terms but the dimension of vector X is preserved.
\begin{itemize}
\item Matrix representation of Encoder and Decoder of PCA are
\end{itemize}
\paragraph{}
$\left[ \begin{array}{c}  x_1 \\ | \\ x_m \end{array} \right] \rightarrow \left[ \begin{array}{c} \ [\vec{q}_1]{^T} \\| \\ \ [\vec{q}_l]{^T} \end{array} \right] \rightarrow \left[ \begin{array}{c} a_1 \\ | \\ a_l \end{array} \right]$
\paragraph{}
$ \left[ \begin{array}{c} a_1 \\| \\a_l \end{array} \right]\rightarrow\left[ \begin{array}{cc} \vec{q}_1 &-- \vec{q}_l \\ \end{array} \right] \rightarrow\left[ \begin{array}{c}\hat{x_1} \\| \\ \hat{x_m} \end{array} \right]$
\begin{equation}
\vec{e} =X-\hat{X}=\sum_{i=l+1}^m a_j \vec{q}_j
\label{equation:eq18}
\end{equation}
Error is found to be orthogonal to $\hat{X}$ , since PCA transforms linear input space into an orthogonal space $\vec{e}.\hat{x} =0$
\begin{itemize}
\item Total variance of m-components of input data $\vec X$
\begin{equation}
\sum_{j=1}^m \sigma_j{^2}=\sum_{j=1}^m\lambda_j
\label{equation:eq19}
\end{equation}
\item Total variance of l -components of $\hat{X}$
\begin{equation}
\sum_{j=1}^l \sigma_j{^2}=\sum_{j=1}^l\lambda_j
\label{equation:eq20}
\end{equation}
\item Total variance of (l-m) components of input data vector in error vector
\begin{equation}
\sum_{j=l+1}^m \sigma_j{^2}=\sum_{j=l+1}^m\lambda_j
\label{equation:eq21}
\end{equation}
\end{itemize}
Where$\lambda_{l+1},\lambda_{l+2},.... \lambda_{m}$ corresponds to small variance. So, the variance of input vectors which are detected by PCA depends on eigen values.
\newline In this section all the equations from 2.8-2.29 are derived 
\paragraph{}$\lambda_i$ are sorted in descending order and the proportion of variance has been explained by principle components(l) as \[\left. \frac {\lambda_1+\lambda_2+.......... +\lambda_l}{\lambda_1+\lambda_2+.....+\lambda_l+..... +\lambda_m}\right.\]If the dimensions are highly correlated, then there will be less number of eigenvectors with large eigenvalues so it has l$\ll$m dimensions and a large dimension reduction may be obtained. We found such typical cases in image and speech processing tasks where inputs are highly correlated. If dimensions are not correlated, then there will be no gain through PCA.
\section{Supervised Learning Method}
Kernel based learning methods are classified into supervised and unsupervised learning algorithms. Kernel method solves any problem by mapping the input data set into high dimensional feature space via linear or nonlinear mapping which is also referred as kernel trick. In recent years, few powerful kernel based models were proposed such as support vector machines, kernel fisher discriminant and kernel principal component analysis which are used for regression, classification, dimensionality reduction and other jobs. In our research work, we adapted Support Vector Machines(SVM) algorithm for regression analysis.
\subsection{Support vector Machines}
Support Vector machine(SVM) is a supervised and powerful learning Based algorithm invented by Vladimir VaDnik~\cite{9233} and it is commonly used for classification and regression analysis. Its formulation is based on structural risk minimization principle which includes capacity control in order to prevent over-fitting problem of Empirical Risk Minimization principle based learning algorithms like traditional Neural Networks. In our research work, we performed regression analysis where Support Vector Regression exhibits the benefits of machine learning with the capability of learning difficult data patterns by mapping of simplified input features extracted from h.264 bit stream data and regressing with desire or true values in a very effective way. The mechanism of SVM works by mapping of nonlinear input data to high dimensional kernel induce space via nonlinear mapping which leads to solving set of linear equations in kernel space~\cite{1410382}. An insensitive loss function is introduced in SVM which measures the risks and the kernel functions has the flexibility that allows SVM to search a wide variety of hypothesis spaces~\cite{1167494}.

\subsection{Regression Analysis}
Support Vector Regression model will provide better accuracy for predicting the video quality in no-reference video quality assessment and line up with human visual system. Support Vector Regression is a supervised learning method in which the input data is mapped into high dimensional kernel induced feature space via nonlinear mapping. Thus a non-linear function is transformed to linear function where regression is performed in high dimensional space. The capacity of the model is controlled by parameters that do not depend on dimension of feature space. In support vector regression, a loss function is introduced called as epsilon that ignores errors situated within the zone of the true value as shown in figure.
\paragraph{}The given input data set X is first mapped onto an m-dimensional feature space using nonlinear mapping, and then a linear model is constructed in the m-dimensional feature space. The linear model in the feature space is given by
\begin{equation}
f(x, \omega)= \sum_{j=1}^{m}\omega_{j}g_{j}(x)+b
\label{equation:eq27}
\end{equation}
$g_{j}(x),j=1,...m$ is a function of nonlinear transformations, and the term b is the bias. Since by pre-processing the input data it is assumed to have zero mean, therefore bias is dropped. The epsilon band with slack variables is shown in the below figure.

Slack variables measure the cost of the errors on the training points and epsilon error is zero for all points that are inside the epsilon band. Support vector regression introduces a loss function called $\varepsilon$-insensitive. Quality estimation can be measured with the help of the loss function.
\begin{equation}
L_{\varepsilon}(y, f(x, \omega))=
\begin{cases}
0, &\text{if $\left| y- f(x, \omega) \right|\leq \varepsilon,$}\\
\left| y- f(x, \omega) \right|-\varepsilon, &\text{otherwise.}
\end{cases}
\label{equation:eq28}
\end{equation}
The empirical risk is defined as
\begin{equation}
R_{emp}(\omega)= \frac{1}{n}\sum_{i=1}^{n}L_{\varepsilon}(y_i, f(x_i, \omega))
\label{equation:eq29}
\end{equation}
The linear regression is performed by support vector regression in higher dimension kernel induced feature space using $\varepsilon$-insensitive loss, also by minimizing tries to reduce complexity of the model. Outside the $\varepsilon$-insensitive zone the deviation of training data samples is measured by introducing slack variables ($\xi_i,\xi_i^{\ast}$  for i=1,....,n) slack variables are non-negative which determines upper and lower bound. Therefore support vector regression is formulated for minimization.
\begin{equation}
min \frac{1}{2}\left|| \omega \right||\ {^2}+C\sum_{i=1}^{n}(\xi_i+\xi_i^{\ast}\ )
\end{equation}
\[
s.t
\begin{cases}
\ y_i-f(x_i, \omega) \leq\varepsilon+\xi_i^{\ast}\\
\ f(x_i, \omega) –y_i\leq\varepsilon+\xi_i\\
\xi_i,\xi_i^{\ast}\geq0, &\text{ i=1,....,n.}
\end{cases}
\]
The optimization can be transformed to the dual problem
\begin{equation}
f(x)=\sum_{i=1}^{n_{sv}}({\alpha-\alpha_i^{\ast}})K(x_i, x))
\end{equation}
\[ w.r.t 0\leq\alpha_i^{\ast}\leq C, 0\leq\alpha_i\leq C\]
\paragraph{}where $n_{sv}$- No of Support Vectors, $\alpha, \alpha_i^{\ast}$ are lagrangian coefficients of certain samples. Some of them  are non zero mean and corresponding samples are support vectors and the kernel function is
\begin{equation}
K(x,x_i)=\sum_{j=1}^{m}g_j(x)g_j(x_i)
\end{equation}
Commonly used kernel functions are: linear kernel function, polynomial kernel function, RBF kernel function. In our proposed model we use RBF as kernel function
\begin{equation}
K(y, y_i)=exp{\frac{-\left||y-y_i \right||{^2}}{\sigma{^2}}}
\end{equation}
In this section the equations from 2.41-2.47 are derived. Where $\sigma$ is kernel parameter.
The generalization performance of support vector regression depends on a good setting of kernel and meta-parameters (C, $\varepsilon$). Optimal parameter selection problem is also complicated by the fact that complexity of SVR model depends on all three parameters. SVM treats meta-parameters as user-defined inputs for regression. Selection of kernel type and kernel function parameters depends on application-domain knowledge and also should reflect distribution of input (x) training data. C determines the trade-off between the model complexity and the amount up to which deviations larger than $\varepsilon$ are tolerated. If C is too large or infinity, then the objective is to minimize the empirical risk, without regard of model complexity part in the optimization formulation . The $\varepsilon$ Controls $\varepsilon$ -insensitive zone width, to fit the training data. The value of this parameter will affect the number of support vectors which are used to construct the regression function, if $\varepsilon$ is large only few support vectors are selected and larger $\varepsilon$-values results in more flat estimates. Hence, both C and $\varepsilon$-values affect complexity of model but in a different way.

\section{Test Methodology}
We preferred k-fold cross-validation (CV) strategy to train and validate the proposed model and Cross Validation(CV) is a technique which was used for estimating performance of predictive model. generally CV is a re-sampling strategy used to validate the performance of our proposed model by random sub-sampling of the available data. In other words, the original data is subdivided randomly into k folds. In each round of CV (k – 1) folds are used for training and the remaining fold is used for validation, this procedure is repeated for k times with each of the k folds should used exactly once for the model validation.  Performance estimation of our proposed model is determined by average of results obtained in k folds for k rounds [5]. Errors which occur on the validation sets are monitored during training. The validation error generally decreases during the initial phase of training, as does the training set error. 
\paragraph{}Mainly, input features are obtained by extraction of bit steam data at decoder side after transmission of video within the network. Our proposed Model has been trained and tested with input features(X) consists of 6 parameters extracted out of bit-stream data as mentioned in above with respective corresponding Target values of SSIM quality metric. 
\begin{table}[ht!]
\begin{tabular}{lllll}
\cline{1-1}\hline
\multicolumn{1}{|l|}{No of instances} & MAE & MSE & RMSE & RSquared \\ \cline{1-1}\hline
Instance 1                            & 0.033981   & 0.002152   & 0.046385    & 0.460293          \\ \hline
Instance 2                            & 0.030235   & 0.001535   & 0.039183    & 0.614883          \\ \hline
Instance 3                            & 0.021469   & 0.000859   & 0.029302    & 0.784624          \\ \hline
Instance 4                            & 0.036298   & 0.002618   & 0.051165    & 0.343333         \\ \hline
\end{tabular}
\end{table}
\paragraph{}The statistical analysis of proposed model explores the values of mean absolute error, mean square error, root mean square error and r square(goodness-of-fit measure) for 4 instances as mentioned in table, quantifying the strength of the relationship between proposed model and the SSIM metric is based on R squared.

\begin{figure}[h]
    \includegraphics[width=1\linewidth]{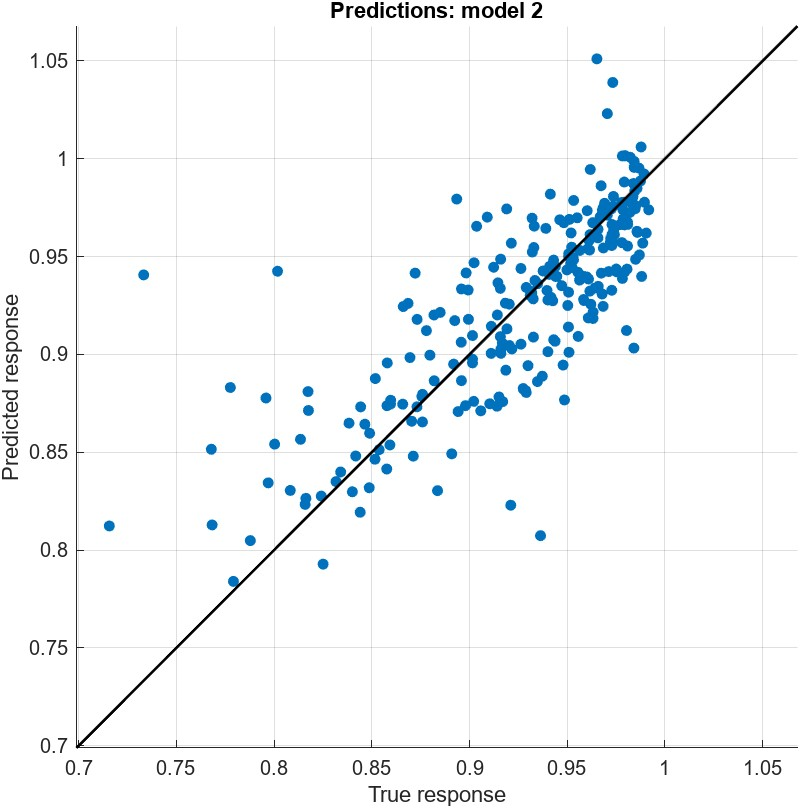}
    \caption{}
  \end{figure}

The Above plot is illustrating linear fit of input data within high dimensional kernel induced space based on support vector machines. 
%\section*{References}

\section{Conclusion}
We concluded that our proposed model exhibited sparseness even after dimension reduction for objective scores of SSIM quality metric. Our Future work is based on improving prediction accuracy which depends on decision towards eliminating sparseness of proposed model.

%\end{thebibliography}

\bibliographystyle{IEEEtran}
\bibliography{refs}
\vspace{12pt}
%\color{red}

\end{document}